\documentclass{article}

\usepackage{microtype}
\usepackage{graphicx}
\usepackage{subcaption}
\usepackage{booktabs} 
\usepackage{enumitem}

\usepackage{hyperref}

\usepackage[preprint]{icml2026}

\usepackage{amsmath}
\usepackage{amssymb}
\usepackage{mathtools}
\usepackage{amsthm}

\usepackage[capitalize,noabbrev]{cleveref}

\theoremstyle{plain}

\theoremstyle{definition}

\theoremstyle{remark}

\usepackage[textsize=tiny]{todonotes}

\icmltitlerunning{Testing the Test: Score-Direction Instability in Class-Split Anomaly Detection}

\begin{document}

\twocolumn[
  \icmltitle{Testing the Test: Score-Direction Instability in Class-Split Anomaly Detection}



  \icmlsetsymbol{equal}{*}

  \begin{icmlauthorlist}
    \icmlauthor{Alejandro Asc\'arate}{sch}
    \icmlauthor{L\'eo Lebrat}{sch}
    \icmlauthor{Rodrigo Santa Cruz}{sch}
    \icmlauthor{Clinton Fookes}{sch}
    \icmlauthor{Olivier Salvado}{sch}
  \end{icmlauthorlist}

  \icmlaffiliation{sch}{School of Electrical Engineering and Robotics, Faculty of Engineering, Queensland University of Technology, Brisbane, Queensland, Australia}

  \icmlcorrespondingauthor{Alejandro Asc\'arate}{a.ascaratecastro@hdr.qut.edu.au}

  \icmlkeywords{Anomaly detection, OOD detection, class-split benchmarks, AUROC inversion, score-direction instability, neighborhood leakage, benchmark diagnostics}

  \vskip 0.3in
]



\printAffiliationsAndNotice{}  

\begin{abstract}
Within-dataset class-split evaluation is widely used as a proxy for fully unconditional out-of-distribution anomaly detection. We show that this protocol can become ill-posed when the held-out anomaly class overlaps the normal mixture in representation space. In this regime, anomaly scores may collapse toward chance or even invert, and the preferred score direction can depend on the unknown anomaly class. We introduce a simple training-free diagnostic, neighborhood class leakage, and show that it predicts score-direction instability across Fashion-MNIST, CIFAR-10, and Imagenette, in both pixel and VAE latent spaces. Our results suggest that class-split AD benchmarks should be treated as geometry-dependent stress tests rather than unconditional evidence of anomaly-detection ability.
\end{abstract}

\section{Introduction} 
A common evaluation strategy in anomaly detection (AD) \citep{chalapathy2019survey,pang2021deepadreview} is to \emph{simulate} anomalies by repurposing existing labeled datasets.
Two families of protocols dominate practice.
First, \emph{cross-dataset} evaluations train on one dataset and treat samples from a different dataset as anomalies, e.g., CIFAR vs.\ SVHN \citep{krizhevsky2009cifar,netzer2011svhn}.
Second, \emph{within-dataset class-split} evaluations designate some semantic classes as normal and others as anomalous (e.g., ``$9$ normal classes vs.\ $1$ anomalous class'').
This protocol is attractive because it appears ``closer'' to fully unconditional-OOD AD: anomalies are not imported from an external source, and the semantic boundary is less contrived.
However, as we show, in the fully unconditional-OOD regime this class-split protocol can become \emph{ill-posed} in a precise operational sense.

The core issue is distributional overlap.
In many natural image datasets, the set of samples from a designated ``anomaly class'' can lie \emph{closer} to the bulk of the (multi-class) normal mixture than a substantial fraction of the normal samples themselves.
In this regime, AD scores that are monotone in distance-to-normality can become unstable across anomaly choices: AUROC can collapse toward chance and may even \emph{invert} (AUROC$<0.5$) for certain anomaly classes.
A tempting reaction is to treat inversion as a cosmetic artifact---e.g., by flipping the score direction---or to argue that a sufficiently strong detector will ``re-invert'' the ranking and restore AUROC.
We argue this misses the underlying failure mode: when the direction of ``anomalousness'' depends on the (unknown) anomaly type, low scores are no longer interpretable in a consistent way.
Consequently, benchmark conclusions become fragile and can reward methods that exploit dataset- and split-specific quirks rather than reflecting robust unconditional-OOD AD capability.

This paper develops a practical diagnostic and an empirical characterization of this phenomenon.
We view class-split AD benchmarks through the lens of \emph{overlap} between class-conditional manifolds in a given representation space.
We introduce a simple, training-free \emph{ill-posedness index} based on local neighborhood class leakage, which quantifies the extent to which class labels fail to align with geometry.
Across datasets and representations, we show that this index predicts when class-split fully unconditional-OOD AD is likely to yield unstable or inverted rankings.

\paragraph{Contributions.}
Our contributions are:
\begin{itemize}
  \item We identify and formalize a practical failure mode of within-dataset class-split evaluation for fully unconditional-OOD anomaly detection: class-dependent score-direction instability, AUROC collapse, and inversion driven by normal--anomaly overlap.
  \item We propose a simple, training-free diagnostic (a neighborhood leakage-based ill-posedness index) that predicts when class-split benchmarking is unreliable in a given representation space.
  \item We empirically validate the phenomenon and the diagnostic across a controlled matrix of settings: three datasets spanning increasing visual complexity (Fashion-MNIST as a negative control, CIFAR-10, and Imagenette), two representations (pixel space and a fully unconditional-OOD VAE latent space), and multiple scoring families (kNN, Isolation Forest, and Local Outlier Factor).
\end{itemize}

\section{Problem setup and evaluation protocol}
Consider a labeled dataset with classes \(y\in\{1,\ldots,K\}\). In a within-dataset class-split protocol, a subset \(N\) is designated normal and a disjoint subset \(A\) anomalous. An AD method is fit only on unlabeled samples from \(D_N\), then evaluated by scoring samples from \(D_N\cup D_A\). We focus on the common \(K-1\) vs. \(1\) protocol: each class \(c\) is held out in turn as the anomalous class, producing class-wise AUROC scores \(\mathrm{AUC}(c)\) \citep{fawcett2006roc}. Class labels are used only for constructing evaluation splits, computing metrics, and computing the diagnostic below; no class-dependent tuning or representation learning is performed.

\paragraph{Hypothesis-test view.}
We interpret the class-split protocol as an implicit test of whether a
fixed representation and scoring rule induce a stable ordering in which
the held-out class is more atypical than the normal mixture. Under this
view, low AUROC is not the only failure mode. A more structural failure
occurs when the preferred score direction varies with the held-out class:
then the benchmark no longer tests a single class-agnostic notion of
anomalousness. The leakage diagnostic below is intended to test this
precondition before class-split AUROC values are used as evidence for
unconditional-OOD AD ability.

\section{A diagnostic for ill-posed class-split benchmarking}
\label{sec:diagnostic}

The within-dataset class-split protocol implicitly assumes that the designated anomalous class is sufficiently separated from the normal mixture in the chosen representation space.
When this assumption fails---i.e., when class-conditional manifolds overlap strongly---the benchmark can exhibit AUROC collapse and class-dependent score-direction instability.
We now introduce a simple, training-free diagnostic that quantifies this overlap and predicts when class-split fully unconditional-OOD AD evaluation is likely to be unreliable.

\paragraph{Representation space.}
Let $r:\mathcal{X}\to\mathbb{R}^d$ denote a fixed representation map.
In our experiments, $r(x)$ is either the vectorized input (pixel space) or a latent code produced by a VAE encoder trained fully unconditional-OOD on normal data.
All definitions below apply to any fixed representation.

\paragraph{Neighborhood class leakage.}
Given a labeled evaluation set $\mathcal{T}=\{(x_i,y_i)\}_{i=1}^m$ with $y_i\in\{1,\dots,K\}$ used \emph{only} for analysis,
let $\mathcal{N}_k(i)$ denote the indices of the $k$ nearest neighbors of $r(x_i)$ among $\{r(x_j)\}_{j\neq i}$ under Euclidean distance.\footnote{Other metrics can be used; we fix Euclidean distance for concreteness.}
We define the \emph{$k$NN class leakage} of a sample $i$ as the fraction of its neighbors whose label differs from its own:
\begin{equation}
\ell_k(i)
\;=\;
\frac{1}{k}\sum_{j\in\mathcal{N}_k(i)} \mathbb{I}\!\left[y_j\neq y_i\right].
\label{eq:leakage-point}
\end{equation}
Intuitively, $\ell_k(i)$ measures the extent to which local geometry in the representation space aligns with semantic class structure:
$\ell_k(i)\approx 0$ indicates that the neighborhood of $x_i$ is class-pure, whereas $\ell_k(i)\approx 1$ indicates heavy mixing with other classes.

\paragraph{Dataset-level ill-posedness index.}
We aggregate pointwise leakage into a dataset-level diagnostic by averaging:
\begin{equation}
L_k(\mathcal{T};r)
\;=\;
\frac{1}{m}\sum_{i=1}^m \ell_k(i).
\label{eq:leakage-index}
\end{equation}
We refer to $L_k$ as an \emph{ill-posedness index} for class-split benchmarking in representation space $r$.
High values of $L_k$ indicate strong overlap of class-conditional manifolds and suggest that holding out a semantic class as ``anomalous'' may not induce a clean normal--anomaly separation.

\paragraph{Why leakage predicts instability.}
Most classical unconditional-OOD AD scores in a representation space $r(x)$ are monotone in either
(i) distance to the empirical support of the normal training set, or
(ii) local density relative to a neighborhood estimate.
If the designated anomalous class occupies regions of the representation space that are densely populated by normal samples (high local mixing),
then many anomalous points will receive low anomaly scores, while some normal tail points may receive higher scores.
As a consequence, the ranking induced by $s(x)$ becomes sensitive to the choice of held-out class and can flip direction for certain anomaly classes.
The leakage index in Eq.~\eqref{eq:leakage-index} provides a direct, training-free proxy for this overlap and thus for benchmark ill-posedness.

\section{Instability and directionality metrics for class-split AD}
\label{sec:instability}

We now define summary statistics that quantify when within-dataset class-split evaluation is unreliable as a proxy for fully unconditional-OOD anomaly detection.
Throughout, fix a dataset with $K$ semantic classes and consider the ``$K\!-\!1$ vs.\ $1$'' protocol where the anomalous class is a single held-out class $c\in\{1,\dots,K\}$.
For a fixed representation $r$ and scoring function $s$, let $\mathrm{AUC}(c)\in[0,1]$ denote the AUROC obtained when class $c$ is treated as anomalous and the remaining $K-1$ classes form the (unlabeled) normal training set.

\paragraph{AUROC dispersion.}
We report the mean, variance, and interquartile range of \(\mathrm{AUC}(c)\) across held-out anomaly classes. High dispersion indicates that the benchmark is sensitive to the arbitrary choice of anomalous class.

\paragraph{Near-random rate.}
To measure how often the protocol yields effectively chance-level rankings, we define the near-random rate at tolerance $\epsilon>0$ as
\begin{equation}
\rho_{\mathrm{rnd}}(\epsilon)
\;=\;
\frac{1}{K}\sum_{c=1}^K \mathbb{I}\!\left[\left|\mathrm{AUC}(c)-\tfrac{1}{2}\right|\le \epsilon\right].
\label{eq:near-random-rate}
\end{equation}
In experiments we use a small fixed tolerance (e.g., $\epsilon=0.05$) to quantify the prevalence of ambiguous rankings.

\paragraph{Inversion rate.}
A stronger failure mode occurs when the AUROC is \emph{inverted}:
anomalous samples are ranked as systematically \emph{more normal} than normal samples, yielding $\mathrm{AUC}(c)<0.5$.
We quantify this via the inversion rate
\begin{equation}
\rho_{\mathrm{inv}}
\;=\;
\frac{1}{K}\sum_{c=1}^K \mathbb{I}\!\left[\mathrm{AUC}(c)<\tfrac{1}{2}\right].
\label{eq:inversion-rate}
\end{equation}
While a single inversion can occur by chance on small test sets, persistent inversions concentrated on specific classes are indicative of substantial normal--anomaly overlap in the chosen representation.

\paragraph{Direction instability.}
A common reaction to inversion is to ``flip'' the score direction.
However, if the preferred direction depends on the (unknown) anomaly type, then the benchmark does not define a consistent notion of anomalousness.
We capture this by tracking the sign of the deviation from chance:
\begin{equation}
d(c)
\;=\;
\mathrm{sign}\!\left(\mathrm{AUC}(c)-\tfrac{1}{2}\right)
\in\{-1,0,+1\},
\label{eq:direction-sign}
\end{equation}
with $d(c)=0$ if $\left|\mathrm{AUC}(c)-\tfrac{1}{2}\right|\le\epsilon$.
We then define a direction-instability rate
\begin{equation}
\rho_{\mathrm{dir}}(\epsilon)
=
1-\frac{1}{K}
\max\left\{
\sum_{c=1}^{K}\mathbb{I}[d(c)=+1],
\sum_{c=1}^{K}\mathbb{I}[d(c)=-1]
\right\}.
\label{eq:direction-instability}
\end{equation}
This quantity is close to $0$ when a single score direction is consistently preferred across anomaly classes and increases toward $0.5$ when the benchmark exhibits mixed directions (also, towards $\geq0.5$ if there are cases in which $d(c)=0$, and thus one should also see an increase in $\rho_{\mathrm{rnd}}$).
This statistic operationalizes the intuition that ``re-inverting'' scores does not resolve the underlying ambiguity when different anomaly classes induce different ranking directions.

\section{Experimental setup}
\label{sec:expsetup}

We evaluate three datasets of increasing visual complexity: Fashion-MNIST \citep{xiao2017fashion} as a low-overlap negative control, CIFAR-10 \citep{krizhevsky2009cifar}, and Imagenette \citep{howard2019imagenette,russakovsky2015imagenet}. For each dataset, we run the \(K-1\) vs. \(1\) class-split protocol, sweeping the held-out anomalous class \(c\). We evaluate two representation spaces: pixels and latent codes \(r(x)=E_\phi(x)\) from a VAE \citep{kingma2014vae} with ResNet-like components \citep{he2016resnet}, trained only on the unlabeled normal pool. In each representation, we fit three standard unconditional AD scorers on the normal pool: kNN distance, Isolation Forest \citep{liu2008isolation}, and Local Outlier Factor \citep{breunig2000lof}. Hyperparameters are fixed a priori and are not tuned per anomalous class.

\section{Results}
\label{sec:results}

\begin{figure}[t]
    \centering
    \includegraphics[width=\columnwidth]{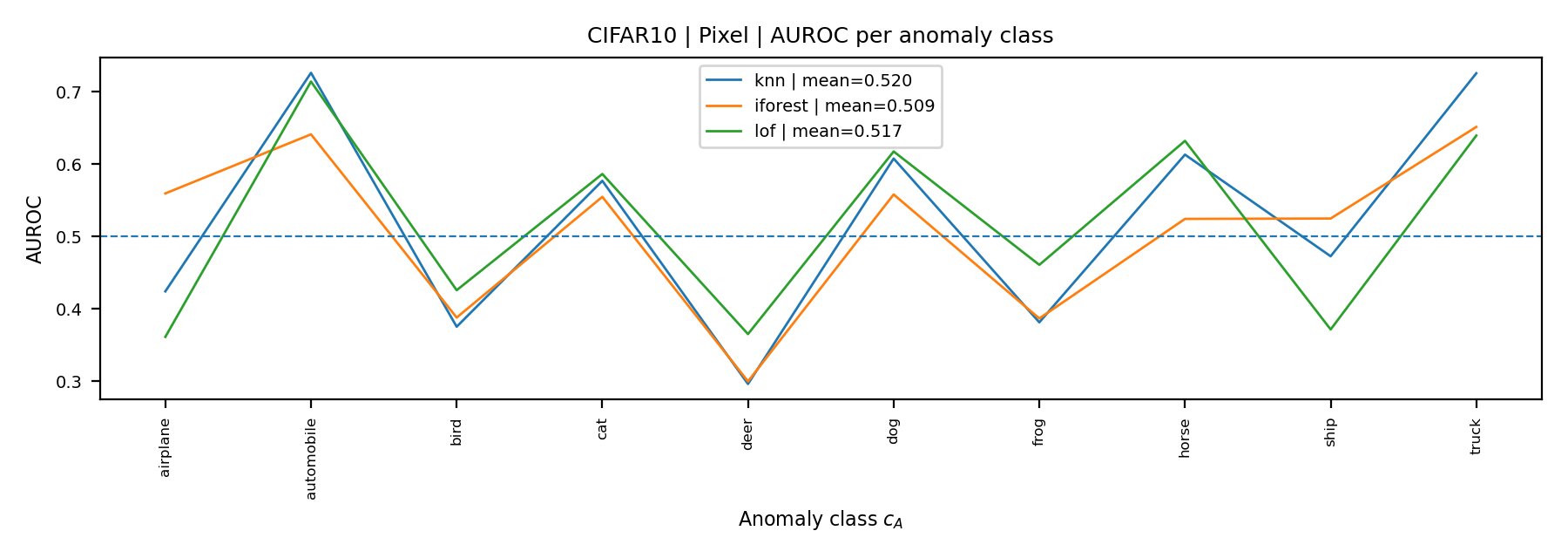}
    \caption{Per-class AUROC on CIFAR-10 in pixel space under the \(9\)-vs.-\(1\) class-split protocol. The dashed line marks chance performance. Different held-out anomaly classes induce both above-chance and inverted rankings, illustrating score-direction instability.}
    \label{fig:cifar10-pixel-auroc}
\end{figure}

Table~\ref{tab:diagnostic} shows the diagnostic relationship between leakage and benchmark instability. Fashion-MNIST has low leakage and comparatively stable class-split behavior, especially in pixel space. By contrast, CIFAR-10 and Imagenette have high leakage in both pixel and VAE latent spaces and exhibit substantial inversion and direction-instability rates. The pattern persists across multiple scoring families after averaging over kNN, Isolation Forest, and LOF, suggesting that the failure is not an artifact of a single detector or representation. Figure~\ref{fig:cifar10-pixel-auroc} shows this effect at the per-class level for CIFAR-10 in pixel space: several anomaly classes lie below chance while others lie substantially above it, despite using the same score convention.

\paragraph{Interpretation.}
The diagnostic should not be read as a replacement for AUROC, but as a
precondition check on the benchmark itself. Low leakage supports the
interpretation that class-split AUROC is measuring anomaly separation in
the chosen representation. High leakage, together with large
\(\rho_{\mathrm{inv}}\) or \(\rho_{\mathrm{dir}}\), indicates that the
protocol is better viewed as a geometry-dependent stress test than as
evidence of general unconditional-OOD detection performance.

\begin{table*}[t]
\centering
\caption{Diagnostic relationship between neighborhood leakage and benchmark instability. Each row is a dataset--representation pair. Instability summaries are averaged over kNN, Isolation Forest, and LOF.}
\label{tab:diagnostic}
\scriptsize
\setlength{\tabcolsep}{4pt}
\renewcommand{\arraystretch}{0.95}
\begin{tabular}{llccccc}
\toprule
Dataset & Rep. & $L_k$ & $\rho_{\mathrm{inv}}$ & $\rho_{\mathrm{rnd}}$ & $\sigma^2_{\mathrm{AUC}}$ & $\rho_{\mathrm{dir}}$ \\
\midrule
Fashion-MNIST & Pixel  & 0.2428 & 0.03 & 0.07 & 0.0194 & 0.10 \\
Fashion-MNIST & Latent & 0.2346 & 0.23 & 0.23 & 0.0224 & 0.30 \\
CIFAR-10      & Pixel  & 0.7609 & 0.43 & 0.13 & 0.0162 & 0.50 \\
CIFAR-10      & Latent & 0.7885 & 0.50 & 0.03 & 0.0185 & 0.50 \\
Imagenette    & Pixel  & 0.7815 & 0.43 & 0.43 & 0.0068 & 0.63 \\
Imagenette    & Latent & 0.8363 & 0.50 & 0.33 & 0.0092 & 0.63 \\
\bottomrule
\end{tabular}
\end{table*}

\section{Discussion}
\label{sec:discussion}

\paragraph{Why ``just use a better AD method'' does not resolve ill-posed class-split benchmarking.}
\label{sec:discussion-reinvert}

A natural counterargument is that AUROC inversion (AUROC$<0.5$) merely indicates that a given scoring function is ``pointing in the wrong direction,'' and that a sufficiently powerful anomaly detector should recover a standard ordering in which anomalies receive larger scores than normal samples.
In this view, the remedy is simply to use a stronger AD method that \emph{re-inverts} the ranking.
We argue that this interpretation conflates two distinct questions:
(i) whether a particular method can be made to achieve AUROC$>0.5$ on a \emph{specific} class-split instance, and
(ii) whether the \emph{benchmark protocol} provides a reliable basis for claims about fully unconditional-OOD AD.
Our results address (ii).

\paragraph{Inversion reflects overlap, not merely a sign convention.}
In the $K\!-\!1$ vs.\ $1$ protocol, the ``normal'' distribution is itself a heterogeneous mixture over $K-1$ semantic classes, often with substantial intra-class variation (backgrounds, pose, lighting, context).
When the held-out anomalous class overlaps this mixture in the chosen representation space, many anomalous samples can lie in regions of high normal density, while some normal samples occupy low-density tails.
Any score that is monotone in distance-to-normality or local density can then yield unstable rankings, and for certain held-out classes the induced ranking can invert.
This is not a cosmetic issue: it indicates that the benchmark does not instantiate a clean separation between normality and the chosen anomaly class in the underlying geometry.

\paragraph{Re-inverting can be achieved in-sample without yielding a meaningful unconditional-OOD notion of anomalousness.}
Even if a more complex detector yields AUROC$>0.5$ for a given anomaly class, this does not imply that the protocol defines a consistent semantic notion of ``anomaly.''
To see why, note that the anomaly class is \emph{unknown} at deployment time.
If different held-out classes induce different preferred score directions (captured by direction-instability $\rho_{\mathrm{dir}}$), then the benchmark is effectively asking a method to solve multiple incompatible tasks:
for some anomaly types, high scores should indicate anomalies; for others (in inverted regimes), low scores would indicate anomalies if one followed the empirical ordering.
A method that ``fixes'' AUROC on the \textit{labeled} evaluation split can do so by exploiting idiosyncratic correlations between class labels and representation geometry, without learning a class-agnostic notion of atypicality.
In this setting, improving AUROC is not equivalent to resolving the ambiguity of what a low (or high) score means for an unseen anomaly type.

Thus the issue is not merely low AUROC, but the loss of a stable score semantics under changes of the held-out anomaly class.

\paragraph{Scope of the diagnostic.}
The proposed leakage index is not meant to certify that a benchmark is
easy or hard in an absolute sense. Rather, it identifies when the
class-split construction ceases to define a stable ordering problem for
a fixed score convention. Low leakage does not guarantee that a detector
will perform well, but it supports the interpretation that AUROC is
testing separation from the normal mixture. High leakage together with
large inversion or direction-instability rates indicates that the
benchmark is probing overlap geometry rather than a class-agnostic
notion of anomalousness.

\section{Conclusion}
\label{sec:conclusion}

Within-dataset class-split evaluation is not always invalid, but its interpretation depends on the geometry of the chosen representation. We showed that, when a held-out class overlaps the normal mixture, class-split AD can exhibit AUROC collapse, inversion, and score-direction instability across anomaly classes. We introduced neighborhood leakage as a simple training-free diagnostic for this failure mode and showed that it tracks instability across datasets, representations, and scorers. These results suggest that class-split benchmarks should report leakage and direction-instability diagnostics before being used as evidence for fully unconditional-OOD anomaly-detection ability.

\bibliography{references} 
\bibliographystyle{icml2026}

\end{document}